\documentclass[10pt,twocolumn,letterpaper]{article}

\usepackage[pagenumbers]{cvpr} 
\pdfoutput=1
\usepackage{graphicx}
\usepackage{amsmath}
\usepackage{amssymb}
\usepackage{booktabs}

\usepackage[algo2e]{algorithm2e} 

\usepackage{algorithm}  
\usepackage{algorithmic}
\DeclareUnicodeCharacter{FB01}{fi}
\usepackage[capitalize]{cleveref}
\Crefname{section}{Section}{Sections}
\Crefname{table}{Table}{Tables}
\crefname{table}{Tab.}{Tabs.}

\def\cvprPaperID{32} 
\def\confName{CVPR}
\def\confYear{2023}

\begin{document}

\title{SAR2EO: A High-resolution Image Translation Framework with Denoising Enhancement}
\author{Jun Yu, Shenshen Du$^{\dag}$, Guochen Xie, Renjie Lu, Pengwei Li, \\
Zhongpeng Cai, Keda Lu, 
\\
University of Science and Technology of China
\\
}




\maketitle

\begin{abstract}
Synthetic Aperture Radar (SAR) to electro-optical (EO) image translation is a fundamental task in remote sensing that can enrich the dataset by fusing information from different sources. Recently, many methods have been proposed to tackle this task, but they are still difficult to complete the conversion from low-resolution images to high-resolution images. Thus, we propose a framework, SAR2EO, aiming at addressing this challenge. Firstly, to generate high-quality EO images, we adopt the coarse-to-fine generator, multi-scale discriminators, and improved adversarial loss in the pix2pixHD model to increase the synthesis quality. Secondly, we introduce a denoising module to remove the noise in SAR images, which helps to suppress the noise while preserving the structural information of the images. To validate the effectiveness of the proposed framework, we conduct experiments on the dataset of the Multi-modal Aerial View Imagery Challenge (MAVIC), which consists of large-scale SAR and EO image pairs. The experimental results demonstrate the superiority of our proposed framework, and we \textbf{win the first place in the MAVIC held in CVPR PBVS 2023}.
\end{abstract}

\section{Introduction}
\label{sec:intro}

Image translation aims at transferring images between the source domain and the target domain by mixing the content and style in an end-to-end manner. Typically, given an image, the image translation task is required to preserve its content while introducing various styles from images in different domains, which makes the task extremely useful in diverse applications, such as face attribute editing and scene style transferring. In the long run, the image translation task has attracted much research attention and many methods are proposed to tackle this problem. Early methods typically adopt style matrix as a mediator in style transferring and show good performance. But there is still much room for improvement. 

Recent years, generative adversarial network(GAN) based models emerge and gradually dominate the image translation tasks. Many effective methods are proposed in this field and gain much popularity, include pix2pix\cite{pix2pix}, cycleGAN\cite{cycleGAN} and pix2pixHD\cite{pix2pixhd}.These models adopt the encoder-decoder structure, where the encoder encodes the input image into a low-dimensional feature vector and the decoder decodes this feature vector into the target image with desired style. During training, the models use paired data and attempt to learn how to convert the style of input image into the target image style via minimizing the difference between the generated image and the real target image. As the GAN based image translation model is capable of generating photorealistic images in image style transferring, it has been widely used in many daily and entertainment scenes. 

However, different from the daily tasks, the remote sensing image translation task, especially the Synthetic Aperture Radar(SAR) and electro-optical (EO) translation task, is still under exploration. Mechanically, the EO images are collected using electro-optical (EO) sensors by capturing images in the visible spectrum (such as RGB and grayscale images). And the SAR images are reproduced through radar signals and act as a complementation of the EO images in the severe conditions such as heavy fog or lack of visible light. The goal of the SAR2EO translation task is to produce high-quality and high-fidelity EO images with SAR images. However, due to the large gap between the SAR and EO images and the heavy noise of images in remote sensing senario, the translation results are often suboptimal.

In this work, we propose a simple but effective framework based on the pix2pixHD with key improvements. Based on the characteristics of SAR and EO images, we propose a denoising enhancement to suppress noise in SAR images. Compared with pix2pix and some other models, the quality of the generated images has been greatly improved. Finally, our solution shows excellent performance on three evaluation metrics: LPIPS\cite{zhang2018unreasonable}, FVD\cite{unterthiner2019fvd}, and L2 Norm, and ranks the first on the leaderboard of the MAVIC held in the CVPR PBVS 2023 with a final score of 0.09. Our main contributions are as follows:
\begin{enumerate}
\item[(1)] 
    On the basis of pix2pixHD, we proposes an effective translation framework named SAR2EO, which demonstrates the capability of converting SAR images into EO images with high quality.

\item[(2)] 
    We propose a denoising enhancement module, which effectively suppresses noise in SAR images while retaining the structural information of the images.

\item[(3)]
    Our proposed method exhibits outstanding performance, leading to the first-place in the MAVIC held in CVPR PBVS 2023.

\end{enumerate}

\section{Related Work}
\label{sec:formatting}

\subsection{GAN}

The GAN model was first proposed in 2014, and generally includes two types of networks, G and D. G stands for generator, which generates images by taking a random code as input and outputting a fake image generated by a neural network. It provides an example of mapping random values to real data. The other network, D, stands for Discriminator, which is trained to distinguish between real and fake data examples generated by the generator. Its input is the image outputted by G, and then it discriminates whether the image is real or fake.In summary, the GAN\cite{GAN} model consists of a generator and a discriminator, which compete with each other to better achieve the task of generating more output that conforms to the mapping relationship for a given input.

Although GAN\cite{GAN} has provided a direction for image translation tasks, traditional GAN\cite{GAN} has some "obvious" flaws: (1) it lacks user control, meaning that in a cGANs\cite{cGANs_1,cGANs_2}, inputting random noise results in a random image. Random images can deceive the discriminator network and have the same features as real images, but they are often not what we want. (2) It has low resolution and quality issues. The generated images may look good, but when you zoom in, you will find that the details are quite blurry. However, over time, researchers have proposed a series of improved GAN network models, such as cGANs\cite{cGANs_1,cGANs_2}, Wasserstein GAN (WGAN)\cite{wasserstein}, and CycleGAN\cite{cycleGAN}, which greatly improve the performance and stability of GAN networks in image translation tasks.

\subsection{ Image-to-Image Translation}

The task goal of image translation is to learn the mapping relationship between the source domain images and target domain images. Depending on the input dataset, image translation can typically be classified into two types: paired and unpaired.

The purpose of supervised learning methods is to learn the mapping relationship between input and output images by training a set of paired image pairs\cite{hoffman2018cycada,pix2pix,park2019semantic,pix2pixhd,zhu2017toward}, which is usually achieved through filename pairing. For example, the training dataset for a facial translation task needs to include pairs of photographs of the same person in different languages. Paired image translation tasks are usually easier to train than unpaired tasks because there is a definite correspondence between images in the dataset.

In fact, it is difficult, and often impossible, to collect dataset with precise pixel-to-pixel mappings. most of the data that we can obtain in real life cannot be one-to-one correspondence, which poses a great obstacle and difficulty for supervised learning. Unpaired image-to-image translation often maps images between two or more domains\cite{rosales2003unsupervised,yi2017dualgan,kim2017learning,benaim2017one,anoosheh2018combogan,tang2019dual,wang2018mix}, where image instances do not match. However, those models can be affected by unwanted images and cannot concentrate on the most usefull part of image. Therefore, the "explorers" of deep learning strive to find GAN\cite{GAN} to solve this problem.

\begin{figure*}[htbp]
	\setlength{\abovecaptionskip}{0cm}
	\setlength{\belowcaptionskip}{0cm}
	\begin{minipage}[b]{1.0\linewidth}
		\centering
		\centerline{\includegraphics[width=17.5cm]{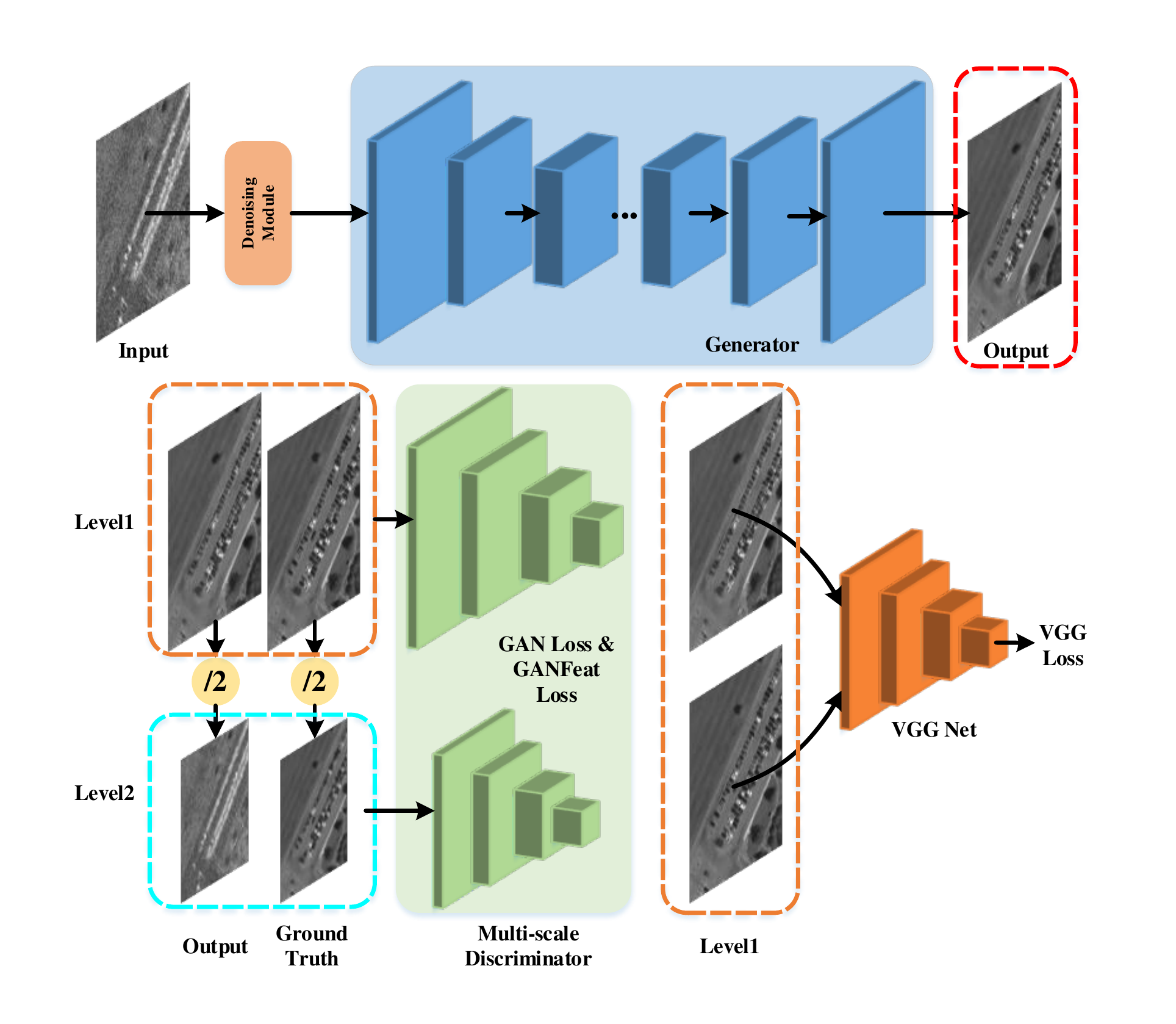}}
	\end{minipage}
	\caption{The overview of our proposed framework.}
	\label{fig:over}
\end{figure*}

\subsection{pix2pix}

The pix2pix\cite{pix2pix} method trains a network by specifying a certain correspondence between input images and output images, imposing constraints on the traditional GAN\cite{GAN} network and no longer allowing it to generate any arbitrary output images that fit the characteristics of real images. In other words, while learning from real images, the pix2pix\cite{pix2pix} network also learns from the input images that can reflect where the real images come from. Pix2pix\cite{pix2pix} ensures that the output image has a certain relationship with the input image, or in other words, given a pair of input and output images, it not only satisfies the characteristics of real images, but also preserves the original information of the input image. This is due to the fact that pix2pix uses the dataset rather than the neural network structure to ensure the correspondence between the output and input, and therefore, the pix2pix network requires paired datasets.

Pix2pix is an "upgrade" to the traditional GAN\cite{GAN}. Instead of inputting random noise, it takes in user-given images and generates images with a structured correspondence to the input. This solves the first drawback of the GAN\cite{GAN} network mentioned earlier.

In the SAR-to-EO challenge, we first conducted experiments using the pix2pix model. Although the training speed of the pix2pix model is relatively fast, the quality of the generated images is poor and the image details are blurry. Meanwhile, the data collected by EO sensors often requires a certain level of clarity. As a result, we ultimately abandoned this approach.

\subsection{CycleGAN}

Pix2pix requires paired data for training, but it is difficult to obtain paired data in real life. CycleGAN, also known as Cycle-constraint Adversarial Network\cite{cycleGAN}, is designed to solve this problem. This type of network does not require paired data (referred to as unpaired datasets), only a set of input data and a set of output data. If pix2pix uses the dataset to ensure the correspondence between the output and input, CycleGAN uses the structure of the neural network to ensure this correspondence. CycleGAN\cite{cycleGAN} adds a restoration network to the GAN network structure, which is used to restore the output and compare the restored image with the original image in terms of pixels to ensure the correspondence between output and input. The success of CycleGAN lies in the separation of style and content.Designing an algorithm to separate style and content manually is difficult, but with neural networks, we can easily change the style while keeping the content constant.

Since the challenge provides paired images, it is convenient for us to conduct supervised training. Therefore, the CycleGAN approach would mean that we are voluntarily giving up this advantage, which we consider unwise.

\section{Proposed Method}
\label{sec:formatting}

This section presents a description of  the proposed translation framework. The framework comprises coarse-to-fine generator, multi-scale discriminators, and a denoising enhancement, as illustrated in \cref{fig:over}. The coarse-to-fine generator is designed to extract both global and local features by combining them. Meanwhile, the multi-scale discriminators are not limited to the input size but are designed to synthesize both the overall image and image details by using smaller and larger discriminators, respectively. The improved adversarial loss further improves the quality of the synthesized images by enhancing their realism. Additionally, the denoising enhancement module is capable of reducing noise in SAR images by applying a non-linear approach to replace the noise.

\begin{figure}[htbp]
  \centering
   \includegraphics[width=1\linewidth]{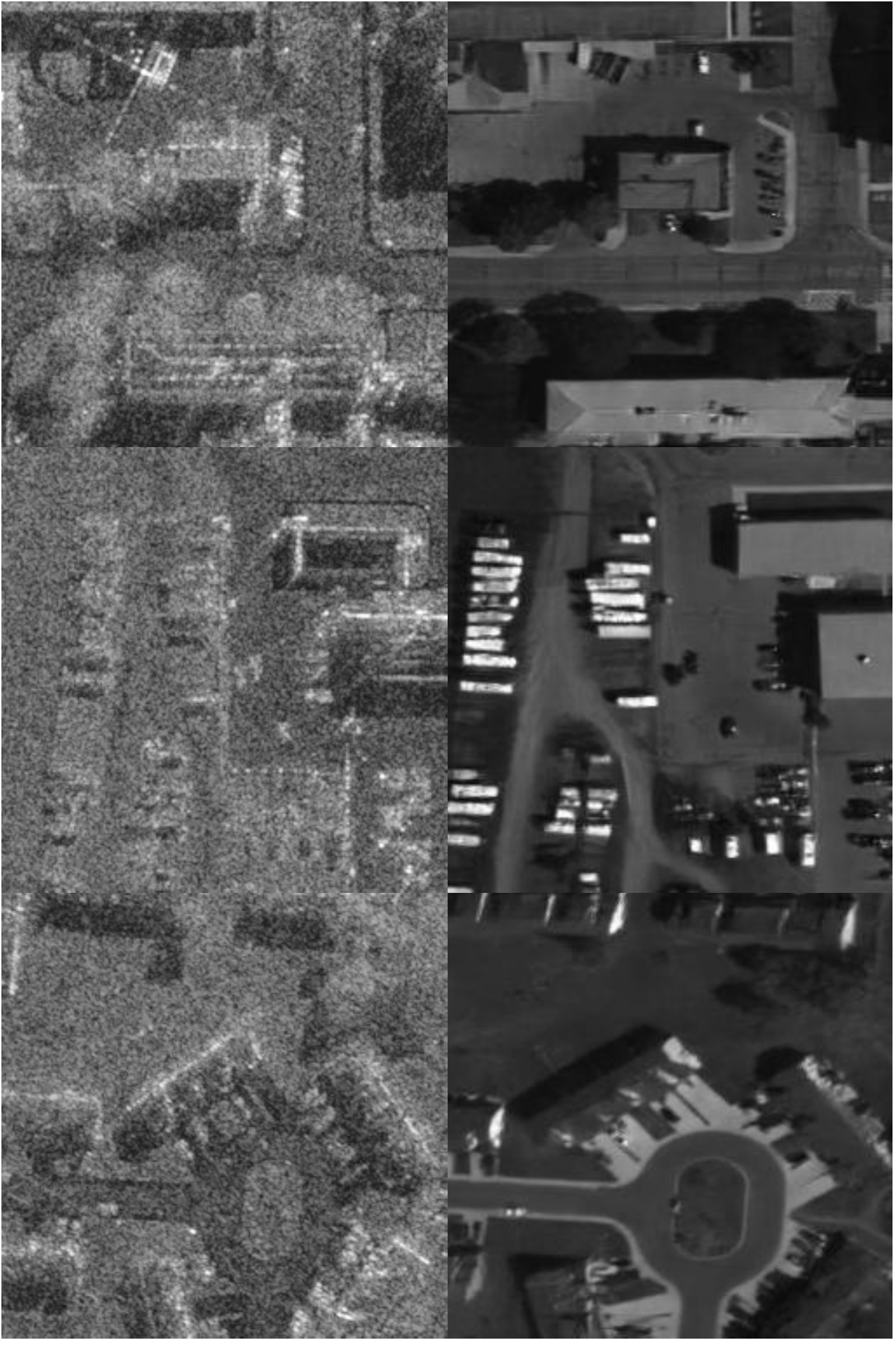}

   \caption{The examples of SAR and EO images.}
   \label{fig:sar_vs_eo}
\end{figure}

\subsection{Preliminary : pix2pixHD}

The pix2pix model is unable to generate high resolution images, and the generated images lack details and realistic texture. Thus, pix2pixHD proposes the following solutions: coarse-to-fine generator, multi-scale discriminators and improved adversarial loss to improve the above problems\cite{pix2pixhd}.

\textbf{Coarse-to-fine generator}: The generator is split into two sub-networks, $G_1$ and $G_2$, with $G_1$ as the global generator and $G_2$ as the local enhancer,  The generator is denoted as $G$ = {$G_1$, $G_2$}, where $G_1$ operates at 1,024 × 512 resolution and $G_2$ outputs an image with 4x the previous output size for synthesizing. $G_2$ can be used to synthesize higher resolution images. $G_1$ is comprised three components: a convolutional front-end $G_1^F$, a set of residual blocks $G_1^R$, and a transposed convolutional back-end $G_1^B$. A semantic label map of resolution 1,024 × 512 is fed through these three components to generate an image of the same resolution.
$G_2$ also has three components: a convolutional front-end $G_2^F$, a set of residual blocks\cite{he2016deep} $G_2^R$, and a transposed convolutional back-end $G_2^B$. The input label map to $G2$ has a resolution of 2,048 × 1,024. The input to $G_2^R$ is the element-wise sum of two feature maps: the output feature map of $G_2^F$ and the last feature map of $G_1^B$. This integrates global information from $G_1$ to $G_2$.
Training involves first training $G_1$, followed by training $G_2$ in order of resolution. Finally, all networks are fine-tuned together. This generator design is effective in aggregating global and local information for image synthesis.

\textbf{Multi-scale discriminators}: For a generative network, designing a discriminator is a rather difficult task. Compared to low-resolution images, for high-resolution images, the discriminator requires a large receptive field, which requires a large convolutional kernel or a deeper network structure. Adding a large convolutional kernel or deepening the network is easy to cause overfitting and will increase the computational burden.
To solve the above problems, multi-scale discriminators have been proposed. It consists of three different discriminators, $D$ = {$D_1, D_2$, $D_3$}, which are the same network structure but operate on different image scales. Then, the generated images are downsampled with a factor of 2 and 4, resulting in three images with different resolutions, which are then inputted into the three identical discriminators. This way, the $D$ corresponding to the image with the smallest resolution will have a larger receptive field, providing a stronger global sense for image generation, while the $D$ corresponding to the image with the largest resolution will capture finer and more detailed features.

\begin{equation}
\begin{aligned}
\mathcal{L}_{\mathrm{FM}}\left(G, D_{k}\right)=\mathbb{E}_{(\mathbf{s}, \mathbf{x})} \sum_{i=1}^{T} \frac{1}{N_{i}} *  \\
\left[\left\|D_{k}^{(i)}(\mathbf{s}, \mathbf{x}) - D_{k}^{(i)}(\mathbf{s}, G(\mathbf{s}))\right\|_{1}\right]
\label{con:fm}
\end{aligned}
\end{equation}

\textbf{Improved adversarial loss}: Since the discriminators have three different sizes and they all are multi-layer convolutional networks, the loss extracts convolutional features from different levels of the synthesized image and matches them with the features extracted from the real image. Then, the feature matching loss is obtained, and its equation is shown \cref{con:fm}:

$D_{k}^{(i)}$ denotes discriminator, where $k$ refers to the $k_{th}$ discriminator and $i$ refers to the number of layers in each discriminator. $N$ refers to the number of elements in each layer. $T$ refers to the number of layers. This feature loss is combined with the GAN's real or fake loss to form the final loss:

\subsection{SAR and EO Images}
SAR is a type of high-resolution imaging radar that can obtain high-resolution radar images similar to optical photographs under extremely low visibility meteorological conditions. SAR image interpretation is very difficult for several reasons. The unique geometric characteristics of SAR images increase the difficulty interpretation\cite{oliver2004understanding}; the inherent coherent speckle noise of SAR images causes target edges to be blurred, and the clarity to decrease, requiring completely different methods for SAR image interpretation. SAR images also have multiple reflection effects, false phenomena, doppler frequency shifts, etc. In contrast, EO images can clearly display imaging edge features, and their resolution is higher than that of SAR images. The difficulty in converting SAR images to EO images lies in eliminating noise in SAR images and improving the resolution of the generated images. \cref{fig:sar_vs_eo} shows the examples of SAR and EO images.

\subsection{Denoising Enhanced SAR2EO Framework}
Based on the characteristics of SAR images and EO images, we propose the SAR2EO solution as shown in the \cref{fig:over}. It is based on pix2pixHD and utilizes the coarse-to-fine generator to fuse global and local features, making the model's feature learning more comprehensive and specific. With the multi-scale discriminators, this method can enhance the details of generated images. The improved adversarial loss improves the quality of the synthesized images by enhancing  their realism. 

\textbf{Denoising enhancement module: }It is common for the energy of signals or images to be concentrated in the low-frequency and mid-frequency bands of the amplitude spectrum, while the information of interest is often submerged by noise in the higher frequency bands\cite{castleman1996digital}. To address the problem of high frequency noise in SAR images, a denoising enhancement module is applied. This module is particularly effective in smoothing noise and can protect sharp edges of the image by choosing appropriate values to replace polluted points. It performs well for salt-and-pepper noise, and is especially useful for speckle noise.

\begin{figure*}[htbp]
	\setlength{\abovecaptionskip}{0cm}
	\setlength{\belowcaptionskip}{0cm}
	\begin{minipage}[b]{1.0\linewidth}
		\centering
		\centerline{\includegraphics[width=17.5cm]{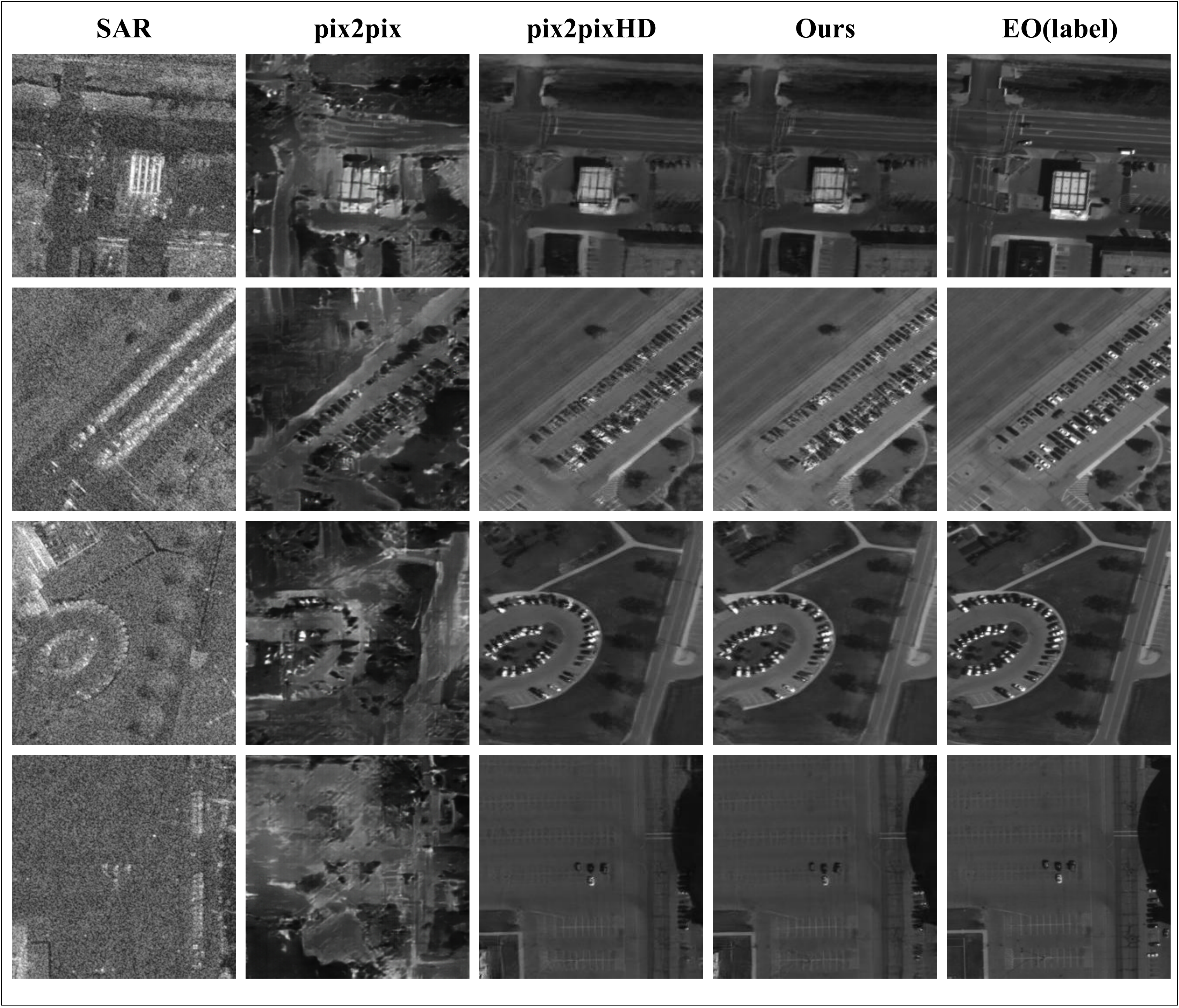}}
	\end{minipage}
	\caption{The results generated by the different models and the corresponding EO images(labels). The first column of images depicts the original SAR images, while the second and third columns showcase the results generated by pix2pix and pix2pixHD, respectively. The fourth column presents the results generated by our framework, and the fifth column displays the corresponding EO images, which serve as the labels for the SAR images. Better viewed in color with zoom-in.}
	\label{fig:five}
\end{figure*}


During the training and inference stages, the SAR images will go through denoising enhancement module to reduce the high-frequency noise of the SAR images and preserve more edge details, and then fed into the generator. As for the EO images used as the label, we hope that the generated images of the model will be closer to this label, so we do no changes to the EO images. This is also the advantage of the image-to-image pair method over the unpaired method. The denoising enhancement algorithm is shown in \cref{alg:al}.





\IncMargin{1em}
\begin{algorithm}[H]
\KwData{An input image $I$ and a window size of $n\times m$.}
\KwResult{An output image $I'$.}
    $I'$ is initialized to $I$;
    \For{$i = 1$ to $height(I)-n$}{
    \For{$j = 1$ to $width(I)-m$}{
    $M$ is a sub-image of size $n\times m$ centered at pixel $(i,j)$;
    $I'_{i,j}$ is set to the median pixel value of $M$;
}
}
\caption{The denoising enhancement algorithm.}
\label{alg:al}
\end{algorithm}
\DecMargin{1em}

\section{Experiments}
In this section, we first discuss the dataset and evaluation metrics briefly, and then introduce the implementation details. We then quantitatively evaluate the performance of our approach on the dataset. Finally, some ablation experiments are conducted to demonstrate the effectiveness of each component.

\subsection{Dataset}
The data for this challenge consists of two types of small window regions (chips) taken from large images captured by several EO and SAR sensors mounted on an airplane. The EO chips are 256 X 256 pix images and belong to targets taken from an airplane. The SAR chips contain roughly the same field of view as the corresponding EO images and are of matching resolution to the EO images. The dataset is divided into: 

\begin{itemize}
  \item [*]
    Train data: This set resembles  the data which is non-uniform and imbalanced.
  \item [*]
    Validation: This set is a uniformly distributed among all classes with about < 100 of samples per class. 
  \item [*]
    Test data: This split resembles the validation test. 
\end{itemize}
The purpose is to use the provided (SAR+EO) train image to design and implement a method for translating SAR images to EO images.

\subsection{Metrics}
It is open and difficult to evaluate the quality of synthesized images\cite{salimans2016improved}. $L$2 Norm, $FVD$ and $LPIPS$ are used as the metrics.

\textbf{L2 Norm}, also known as Euclidean distance or $L2$ distance, is a commonly used distance metric to measure the difference between two vectors.

The $L2$ Norm measures the length of the vector, which is the distance from the origin to the point represented by the vector. In image processing, the $L2$ Norm is often used to calculate the pixel-wise difference between two images.

\textbf{LPIPS} (Learned Perceptual Image Patch Similarity) is a perceptual image quality metric that measures the similarity between two images based on the response of deep neural networks. $LPIPS$ was introduced in a paper by \cite{zhang2018unreasonable} and has been shown to correlate well with human perception of image quality.
The $LPIPS$ metric is calculated by passing two images through a pre-trained deep neural network and computing the distance between the feature representations of the two images. The distance metric used is typically the $L2$ Norm. The final $LPIPS$ score is obtained by averaging the distances over multiple image patches.

\begin{table*}[htbp]
  \centering
  \begin{tabular}{@{}|c|c|c|c|c|c|c|@{}}
  \hline
    User &Final Score &LPIPS &FVD &L2 &CPU[1]/GPU[0] &Extra Data/No Extra Data \\
    \hline
    \textbf{USTC-IAT-United} &\textbf{0.09} &\textbf{0.25} &\textbf{0.02} &\textbf{0.01} &\textbf{0.00} &\textbf{0.00}\\
    \hline
    pokemon &0.14 &0.35  &0.04 &0.01 &-1.00 &-1.00\\
    \hline
    wangzhiyu918 &0.14 &0.38  &0.02 &0.01 &0.00 &0.00\\
    \hline
    ngthien &0.18 &0.43  &0.10 &0.01 &0.00 &0.00\\
    \hline
    Wizard001 &0.26 &0.50  &0.27 &0.02 &-1.00 &-1.00\\
    \hline
    hanhai &0.30 &0.30  &0.59 &0.01 &0.00 &0.00\\
    \hline
    u7355608 &0.33 &0.54  &0.43 &0.02 &-1.00 &-1.00\\
    \hline
    jsyoon &0.33 &0.46  &0.53 &0.01 &0.00 &0.00\\
    \hline
  \end{tabular}
  \caption{PBVS 2023 Multi-modal Aerial View Imagery Challenges - Translation test set leaderboard.}
  \label{tab:1}
\end{table*}

where $I1$ and $I2$ are the two images being compared, $fi$ is the feature extractor for the $i$-th image patch, and $n$ is the total number of patches.

\textbf{FVD} (Fréchet Video Distance) is a metric that measures the similarity between two sets of images. $FVD$ was introduced in a paper by \cite{unterthiner2019fvd} and is based on the distance between the feature representations of the images calculated by a pre-trained deep neural network.
The $FVD$ metric is calculated by first computing the mean and covariance of the feature representations of the real images and the generated images. The distance between the mean and covariance is then computed using the Fréchet distance, which is a measure of similarity between two multivariate Gaussian distributions. A lower $FVD$ score indicates a higher similarity between the two sets of images.


where $mu_real$ and $sigma_real$ are the mean and covariance of the feature representations of the real images, $mu_fake$ and $sigma_fake$ are the mean and covariance of the feature representations of the generated images, and $Tr()$ denotes the trace of a matrix.


The evaluation indicator on the final ranking is the average of the above three indicators.

\begin{figure}[htbp]
  \centering
   \includegraphics[width=1\linewidth]{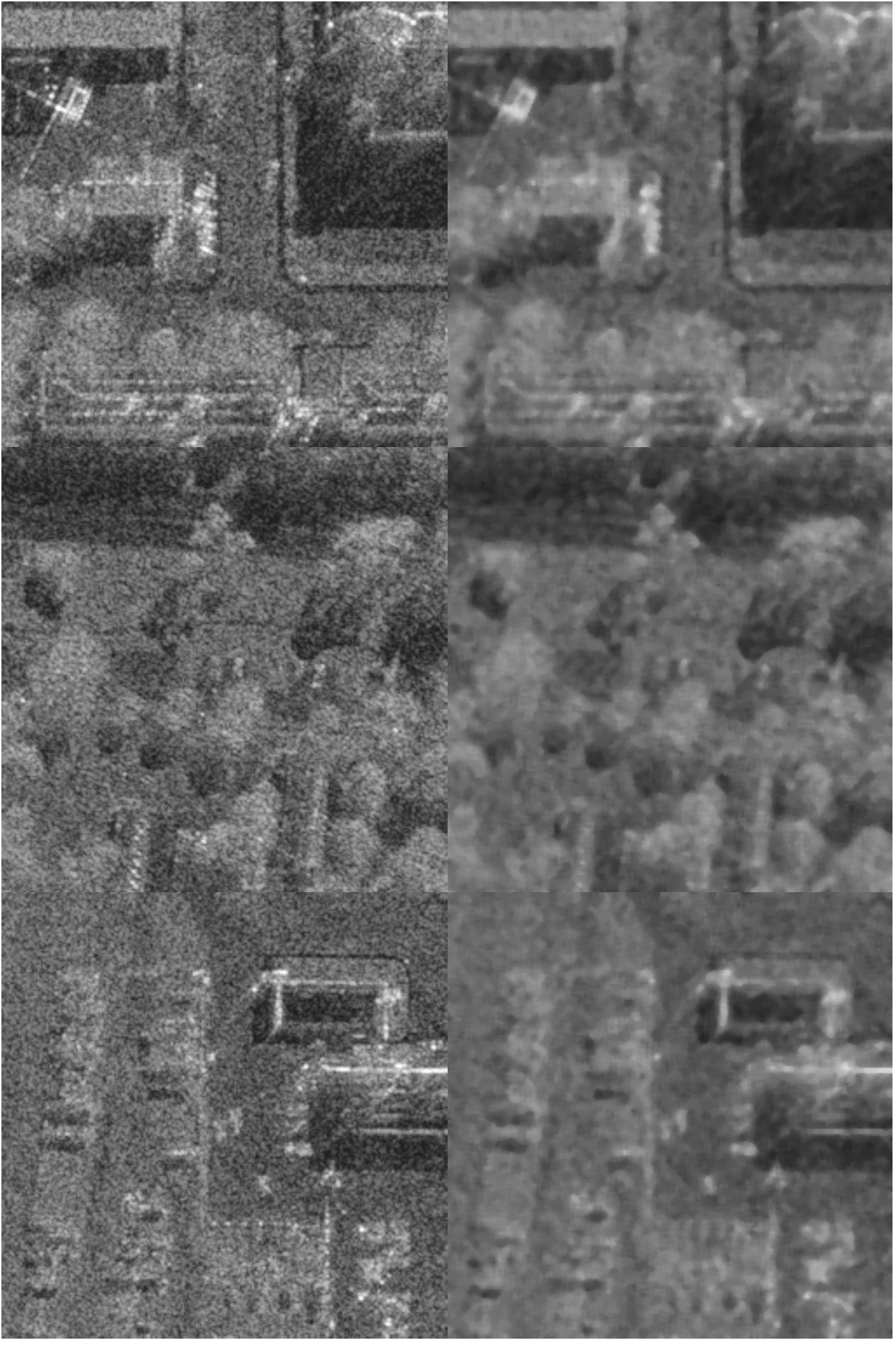}

   \caption{SAR images and images after denoising enhancement.}
   \label{fig:sar_vs_eo}
\end{figure}

\subsection{Implementation Details}
Development phase/learning: Since we have the validation set and the corresponding labels accessible at this phase, we perform the validation offline, train on the training set, and then test on the validation set.

Final Evaluation: In this phase, the training and validation sets were jointly used as the training set to expand the data. The experiments were performed on NVIDIA GPU 3090.

\subsection{Main results}
\textbf{Challenge overview: }Sensor translation algorithms allow for dataset augmentation and allows for the fusion of information from multiple sensors. EO and SAR sensors provide a unique environment for translation. The motivation for this challenge is to understand how if and how data from one modality can be translated to another modality. This competition challenges participants to design methods to translate aligned images from the SAR modality to the EO modality.

In the context of the Multi-modal Aerial View Imagery Challenge (MAVIC), the team USTC-IAT-United attained the first rank in the ultimate test set ranking. This feat was accomplished through the attainment of noteworthy performance metrics, specifically a LPIPS score of 0.25, a FVD score of 0.02, and a L2 score of 0.01. The composite score of the team was computed as 0.09, representing the average of the aforementioned performance indicators. The obtained results provide compelling evidence of the efficacy of the team's competition strategy. The final test set leaderboard is shown in \cref{tab:1}.
Additionally, we compared our proposed method with other image translation models, namely pix2pix and pix2pixHD. The training data provided by the competition was utilized for training, and the validation set was used for evaluation, with LPIPS, FVD and L2 being the chosen metric. The performance of the translators is presented in a \cref{tab:2}.


\begin{table}[htbp]
  \centering
  \begin{tabular}{@{}|c|c|c|c|@{}}
  \hline
    Method &LPIPS &FVD &L2\\
    \hline
    pix2pix &0.484638 &0.08 &0.02  \\
    \hline
    pix2pixhd &0.259611 &0.02 &0.01\\
    \hline
    \textbf{ours} &\textbf{0.253924} &\textbf{0.02} &\textbf{0.01}\\
    \hline
  \end{tabular}
  \caption{The effect of using different schemes on the experimental results, the LPIPS, FVD and L2 metric on the local validation dataset.}
  \label{tab:2}
\end{table}

The images generated by different methods are showcased in \cref{fig:five}. Regarding the quality of the results, SAR images generally exhibit a high degree of noise and blurry object boundaries, leading to a patchy visual appearance. The second column showcases the results generated by the pix2pix model, which suffers from severe distortions, with object shapes being deformed and inaccurate color generation. However, the generated images still exhibit basic shape contours, albeit with blurry details. In contrast, the third column depicts the results generated by pix2pixHD, which exhibit a higher level of image clarity and closely resemble the labels in terms of shape, with improved color accuracy. Nevertheless, some details still require refinement when compared to the label images. Finally, the results produced by our framework exhibit even better quality compared to pix2pixHD, with shapes and colors that are more similar to the label images, and sharper details in the generated images.
\subsection{Ablation studies}
In this section, we conducted ablation experiments to demonstrate the effectiveness of our framework. The validation dataset was utilized for the experimentation. Furthermore, visualizations of the images in this section were presented to illustrate the efficacy of our denoising enhancement.

\textbf{Effectiveness of denoising enhancement module of our framework:} In the \cref{tab:3}, the metric is slightly improved, it shows the effectiveness of denoising enhancement.

\raggedbottom
\begin{table}[htbp]
  \centering
  \begin{tabular}{@{}|c|c|c|c|@{}}
  \hline
    Method &LPIPS &FVD &L2\\
    \hline
    pix2pixHD &0.259611 &0.02 &0.01\\
    \hline
    +\textbf{denoising enhancement} &\textbf{0.253924} &\textbf{0.02} &\textbf{0.01}\\
    \hline
  \end{tabular}
  \caption{Conducting the experiments on the pix2pixHD and pix2pixHD with denoising enhancement, the metric is LPIPS, FVD and L2 on the local validation dataset.}
  \label{tab:3}
\end{table}

\setlength{\parskip}{1pt}

\section{Conclusion}
This paper presents a novel framework for converting SAR images to EO images. Our framework is based on pix2pixHD architecture, which is known for its ability to generate high-quality images. The coarse-to-fine generator, multi-scale discriminators, and improved adversarial loss in pix2pixHD are utilized to enhance the resolution and quality of the generated EO images. To further improve the conversion quality, we analyzed the specific characteristics of SAR and EO images and proposed a denoising enhancement module to reduce the noise in SAR images and enhance the contours of objects in SAR images. This denoising enhancement module was incorporated into the overall framework and resulted in a significant improvement in image quality. The proposed framework has several advantages over existing methods. First, our framework is highly effective in generating high-quality EO images from SAR images, which is a challenging task due to the significant differences between the two modalities. Second, our denoising enhancement module effectively reduces the noise in SAR images, which is a common issue in SAR images. Third, our framework is based on the widely-used pix2pixHD architecture, making it easy to implement and deploy in various applications. We evaluated the performance of our framework in the MAVIC competition, where the goal was to generate high-quality EO images from SAR images. Our framework achieved a final score of 0.09, which was the highest among all participating teams and secured the first place in the competition.


{\small
\bibliographystyle{ieee_fullname}
\bibliography{egbib}
}

\end{document}